\documentclass[10pt,journal,compsoc]{IEEEtran}
\usepackage[accsupp]{axessibility}
\usepackage{times}
\usepackage{epsfig}
\usepackage{graphicx}
\usepackage{svg}
\usepackage{amsmath}
\usepackage{amssymb}
\usepackage{multirow}
\usepackage{array}
\usepackage{tabularx}
\usepackage{caption}
\usepackage{subcaption}
\usepackage{hyperref}
\usepackage{float}
\usepackage{booktabs}
\usepackage{comment}
\usepackage{subcaption}
\usepackage[table]{xcolor}
\hypersetup{
    colorlinks=true,
    linkcolor=blue,
    filecolor=magenta,      
    urlcolor=cyan,
    pdftitle={Overleaf Example},
    pdfpagemode=FullScreen,
    }

\newcolumntype{P}[1]{>{\centering\arraybackslash}p{#1}}

\begin{document}

\title{Fusion-SSAT: Unleashing the Potential of Self-supervised Auxiliary Task by Feature Fusion for Generalized Deepfake Detection}

\author{Shukesh Reddy, Srijan Das, Abhijit Das,~\IEEEmembership{Senior Member ~IEEE}
        
        
\IEEEcompsocitemizethanks{
\IEEEcompsocthanksitem S. Reddy, A. Das are with the Machine Intelligence Group, Department of CS\&IS, Birla Institute of Technology and Sciences, Pilani,
India.\protect\\
S. Das, University of North Carolina at Charlotte, United States of America.\\
E-mail: abhijit.das@hyderabad.bits-pilani.ac.in

}

}

%
%

\markboth{IEEE Transactions on Biometrics, Behavior, and Identity Science,~Vol.~XX, No.~XX, MM~YYYY}%
{S.Reddy \MakeLowercase{\textit{et al.}}: Fusion-SSAT: Unleashing the Potential of
Self-supervised Auxiliary Task by Feature
Fusion for Generalized Deepfake Detection}

\IEEEtitleabstractindextext{
\begin{abstract}
In this work, we attempted to unleash the potential of self-supervised learning as an auxiliary task that can optimise the primary task of generalised deepfake detection. To explore this, we examined different combinations of the training schemes for these tasks that can be most effective. Our findings reveal that fusing the feature representation from self-supervised auxiliary tasks is a powerful feature representation for the problem at hand. Such a representation can leverage the ultimate potential and bring in a unique representation of both the self-supervised and primary tasks, achieving better performance for the primary task. We experimented on a large set of datasets, which includes DF40, FaceForensics++, Celeb-DF, DFD, FaceShifter, UADFV, and our results showed better generalizability on cross-dataset evaluation when compared with current state-of-the-art detectors.
\end{abstract}

\begin{IEEEkeywords}
Generalized Deepfake Detection, Self-supervised Learning, Auxiliary Task, Local Directional pattern. 
\end{IEEEkeywords}

}

\maketitle

\vspace{-24mm}
\section{Introduction}
\IEEEPARstart{R}{apid} increase in the use of Large Language Models (LLMs)\cite{openai2024gpt4technicalreport,deepseekai2025deepseekv3technicalreport}, deep denerative models\cite{open-sora} and diffusion models\cite{rombach2021highresolution} has revolutionized content creation across text, image, video, and audio domains. These advancements enabled the ease in creating of extraordinarily realistic synthetic media, usually indistinguishable from real content. Accelerated growth in deepfake generation techniques\cite{K_2024_CVPR,wang2022latent,ren2021pirenderer,DBLP:conf/mm/ChenCNG20,10.1145/3394171.3413532,choi2018stargan,Patashnik_2021_ICCV,bounareli2023hyperreenact,shiohara2023blendface,ren2023pbidr,an_2021_pfc_iccvw,ho2020denoising,hong2022depth} with respective to Face swapping\cite{shiohara2023blendface,Xu_2022_CVPR}, Facial expressions\cite{an_2021_pfc_iccvw,DBLP:conf/mm/ChenCNG20,choi2018stargan}, face reenactment\cite{bounareli2023hyperreenact,Hsu_2022_CVPR,head2head2020,liu2024tperson}, talking face generation\cite{hong2022depth,10204743,Sha2024ZeroFake,wang2024comparative}, facial attribute editing\cite{dalva2022vecgan,lu2024towards,yang2024fncontra,wang2022fregan} from a source image or video to a target video. This forgery poses a risk to subject identifications that have been widely employed in digital payments, video surveillance, and social media. To mitigate these concerns, there is a rapidly expanding research on deepfake detectors for the identification of facial forgeries. 
As a result, many research efforts have focused on developing methods to detect manipulated media using CNN and RNN-based models: XceptionNet\cite{chollet2017xceptiondeeplearningdepthwise}, EfficientNetB7\cite{tan2020efficientnetrethinkingmodelscaling}; Transformer-Based Models: vision transformers (ViT)\cite{dosovitskiy2021imageworth16x16words}, swin transformer\cite{liu2021swintransformerhierarchicalvision}. 

Although current state-of-the-art deepfake detectors, UCF\cite{ucf}, RECCE\cite{Cao_2022_CVPR}, CORE\cite{Ni_2022_CVPR}, FFD\cite{FFD}, Face X-Raycite\cite{Face-X-Ray}, DSP-FWA\cite{DSP-FWA}, Capsule\cite{capsule}, and XceptionNet\cite{chollet2017xceptiondeeplearningdepthwise} exhibit strong performance in within-domain evaluations, they demonstrate a significant decrease in performance when employed in cross-domain or cross-dataset analyses. 

State-of-the-art models\cite{SimCLR,SRM,Cao_2022_CVPR,capsule,clip,DSP-FWA,FFD,Face-X-Ray,Diet,Dino,HRNet,DBLP:conf/mm/ChenCNG20}often struggle to generalize across datasets, as they tend to learn superficial, dataset-specific patterns rather than learning the underlying generative mechanisms of manipulated content. The over-reliance on dataset-specific artefacts limits their ability to perform reliably in cross-dataset evaluations. The resulting drop in performance reveals a critical shortcoming in the model's ability to learn robust and transferable features that hold up across varied domains.

While these models may achieve high accuracy in controlled or domain-specific settings \cite{ucf,Ni_2022_CVPR}, they frequently overfit to idiosyncratic cues present in the training data, such as compression artefacts or content regularities thereby diminishing their practical utility. In real-world scenarios, where deepfakes vary widely in terms of generation methods, compression levels, and content diversity, such limitations become especially pronounced, underscoring the need for more resilient and generalizable detection approaches.

To mitigate the above-mentioned limitation, we proposed a novel approach, Fusion-SSAT, which consists of an auxiliary task reconstruction and deepfake detection as the main task. The reconstruction task aims to learn the local texture features from a given masked RGB face image. The primary task is a global feature encoder for binary classification of deepfake detection. We fuse the feature from the encoder of the auxiliary task along with the feature from the encoder of the primary detection task to blend both the global and local features. This fusion of local and global features significantly enhances the performance of deepfake detection across datasets, as the reconstruction of local texture patterns from the corresponding RGB pushes the encoder to identify intricate face artefacts and distortions created during manipulation, hence empowering the detection feature representation. To be specific, this significantly more arduous process of featuring is prompted to acquire manipulation resilient representations. Hence, it served as a means of implicit regularisation, pushing the model to more effectively detect local cues along with the global cues that often differ between real and fake content. 

To summarize, our contributions are as follows:
\begin{itemize}
\item A novel approach, Fusion-SSAT, to improve deepfake detection and to generalization in cross-domain deepfake detection scenarios.
\item Demonstration of Fusion-SSAT’s ability to learn fine-grained texture and artefact patterns via local texture pattern from its corresponding face images.
\item Extensive evaluation on DF40 and deepfake benchmark datasets validating the effectiveness of the proposed approach.
\end{itemize}
\section{Related Works}
In deepfake detection, CNN's are frequently preferred over alternative architectural models.  Preliminary research in deepfake detection has primarily utilised known CNN architectures, such exception-net\cite{chollet2017xceptiondeeplearningdepthwise} and efficientnet B4\cite{tan2020efficientnetrethinkingmodelscaling}, as binary classifiers.  Nonetheless, these techniques encounter difficulties in generalizing to unseen manipulation methods.  To tackle this difficulty, various solutions have been explored, including disentanglement learning, multi-task learning\cite{das-limiteddatavit-wacv2024,zou2024semanticsorientedmultitasklearningdeepfake,Pbalaji}, and pseudo fake synthesis\cite{laanet,sbi}.  Advancements in generative modeling have led to deepfakes becoming increasingly realistic, resulting in imperceptible localized errors. The previously described methods are susceptible to obsolescence. They continue to depend on conventional CNNs\cite{chollet2017xceptiondeeplearningdepthwise,tan2020efficientnetrethinkingmodelscaling,Roy2022}, hence experiencing a loss of local information with successive convolutional layers. To enhance the capture of low-level features, techniques employing implicit attention strategies\cite{Diet} have been suggested nevertheless, they exhibit inadequate generalization.

Recent advancements in vision transformers (ViTs)\cite{dosovitskiy2021imageworth16x16words,P1,P2,Pbalaji} have resulted in the creation of multiple types for diverse applications. Optimal performance of these models necessitates comprehensive datasets and pre-training.  The DeiT model\cite{Diet} was improved to reduce these requirements by advanced regularization, data augmentation, and token extraction from convolutional layers.  The tokenisation method utilised by T2T\cite{T2T} was implemented to discern and document local structural information.  Convolutional filters were employed in alternative models to incorporate inductive bias.  Hierarchical transformers\cite{HAT} have incorporated inductive bias by reducing the number of tokens through patch merging.  Nonetheless, they still require datasets of considerable size for pre-training.

SSL has shown effective usage in image-based techniques such as SimCLR\cite{SimCLR}, MoCo\cite{moco}, MAE\cite{MAE}, and DINO\cite{Dino}, as well as in video-based methodologies like CoCLR\cite{coclr}, VideoMAE\cite{videomae}, and SVT\cite{SVT}.  SSL emphasis the acquisition of representations by utilizing the intrinsic structure of data, hence obviating the dependence on labeled data.  This approach has demonstrated considerable enhancements in generalizability, with current research exceeding the performance of supervised models in zero-shot testing.  Leveraging these developments, we expect that ssl will facilitate the model's acquisition of superior representations to improve generalizability in deep fake detection.  The SSL-based method enables the model to accurately differentiate between authentic and counterfeit instances, regardless of the manipulation strategy utilized.

MTL has been extensively studied across multiple domains of machine learning and deep learning applications.  It has been utilized in natural language processing applications, including unified representations and representation learning.  Furthermore, MTL has been utilized in voice recognition, drug discovery, and computer vision applications such as facial analysis, pedestrian identification, facial alignment, and attribute prediction, among others.  Moreover, MTL has recently acquired importance in face attribute learning, sometimes referred to as semantic features, as they offer a more authentic depiction of objects and actions.  This method facilitates a thorough comprehension of the visual domain by concurrently modeling several aspects in face-related activities. Although MTL is frequently examined through Supervised Learning, its investigation within the framework of SSL is still mostly uncharted, rendering it particularly pertinent to the current issue.  Furthermore, it is important to highlight that deep fake detection has not been thoroughly examined within the MTL framework.

\section{Proposed methodology}
\vspace{-1mm}
\subsection{Preliminaries}

\begin{figure*}[h]
    \centering
    \includegraphics[width=0.8\textwidth]{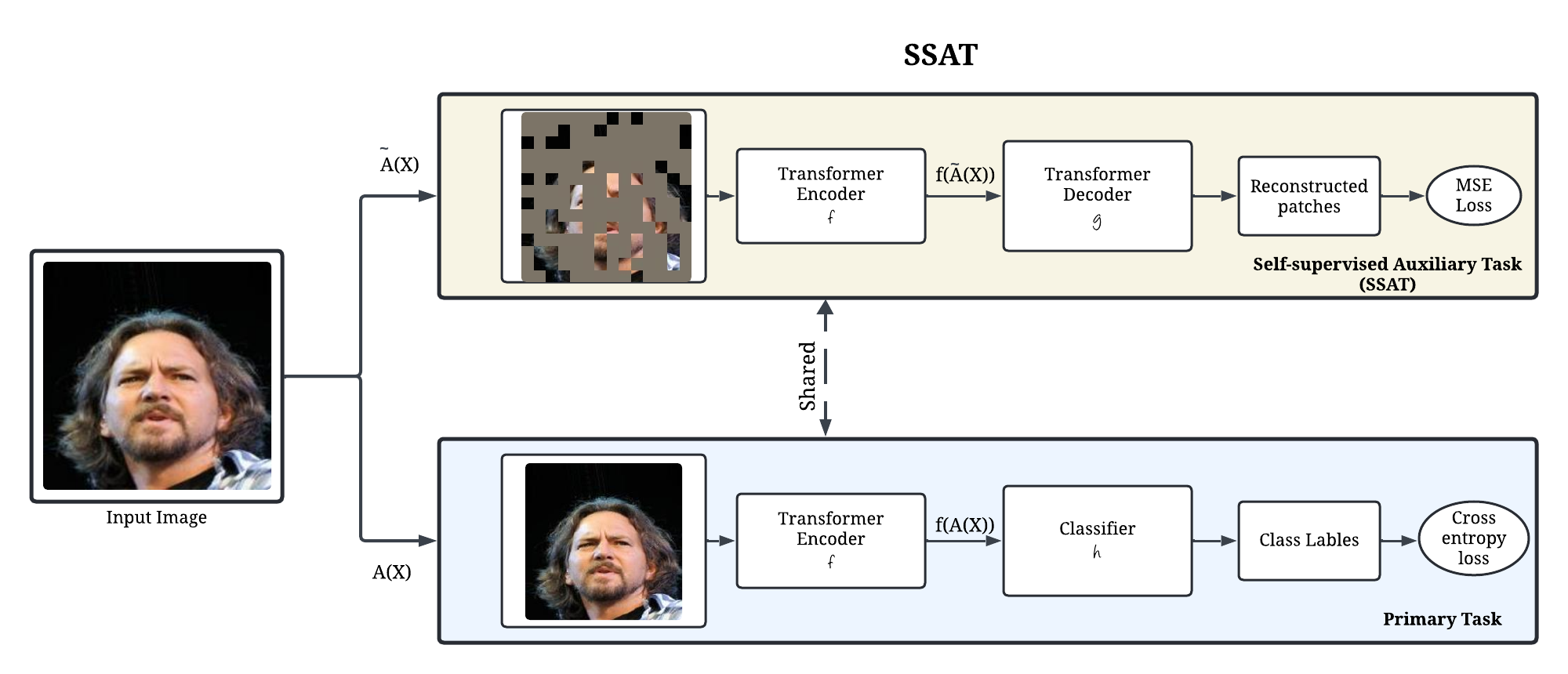} 
    \caption{Overview of the self-supervised auxiliary task (SSAT) framework.}
    \label{fig:ssat}
\end{figure*}

\noindent\textbf{Self-Supervised
Auxiliary Task (SSAT)\cite{das-limiteddatavit-wacv2024}:} Jointly optimizes ViTs for the Primary task (classification) and a self-supervised auxiliary task (reconstruction) when the amount of training is limited.  The primary task utilizes the latent representation of full image for the classification, while masked image is used for the reconstruction task. The framework is jointly trained on losses of primary task denoted by \( L_{\text{cls}} \) and auxiliary task denoted by \(L_{\text{SSAT}} \), where  \(\lambda\) is set to 0.1 thereby assigning a higher relative weight of 0.9 to \(L_{\text{SSAT}} \) during training shown in equation~\ref{ssat_loss}. Figure~\ref{fig:ssat} provides an overview of the framework. 

\begin{equation}
   L = \lambda \cdot L_{\text{cls}} + (1 - \lambda) \cdot L_{\text{SSAT}}
   \label{ssat_loss}
\end{equation}
SSAT showed a better generalizability in high quality deepfakes but fails to learn the local and global features from highly compressed deepfake's. 

\noindent\textbf{Local Texture Features:} encode fine-grained spatial and directional variations in facial regions, offering robustness against illumination changes, noise, and image compression. By analysing edge responses across extended neighbourhoods, these descriptors preserve intricate local structures around critical regions such as the eyes, nose, and mouth. Advanced sampling and directional weighting enhance their ability to represent subtle structural differences that may be overlooked by global descriptors. This enables local texture features to serve as a powerful tool in applications that demand precise discrimination of subtle and fine-grained visual patterns.
   
\subsection{Fusion-SSAT}


In contrast to state-of-the-art deepfake detectors, training solely on classification tasks will limit the ability to understand various patterns of deepfake artifacts.  In Fusion-SSAT, we introduced a multitask learning framework where in RGB video serves as input for the primary task of deep fake detection i.e real/ fake classification, conducted in a supervised manner, while LDP video/ local texture pattern is reconstructed as auxiliary task executed in a self-supervised manner, with shared features between the tasks. To enhance the generalisation of deepfakes, we train the classifier using a combination of features from the encoder of the auxiliary task and global features from RGB, which facilitates improved pattern recognition compared to training only on global features. 

Let \( V = \{v_1, v_2, \dots, v_n\} \) represent a dataset of video sequences, where each \( v_i \) is a video comprising a sequence of frames organized as a tensor \( v_i \in \mathbb{R}^{B \times T \times C \times H \times W} \). Here, \( B \) denotes the batch size, \( T \) is the number of frames, \( C \) is the number of channels, \( H \) is the frame height, and \( W \) is the frame width. Each video is provided in two forms: a masked RGB video \( L(v_i) \) and an full patch RGB video \( R(v_i) \), both with dimensions \( \mathbb{R}^{B \times T \times C \times H \times W} \). The RGB video undergoes random masking with a masking ratio of 0.75, denoted as \( \tilde{L}(v_i) = M(L(v_i)) \), where \( M \) randomly masks 75\% of the spatio-temporal patches. 

Both \( \tilde{L}(v_i) \) and \( R(v_i) \) are processed by a shared ViT encoder \( f : \mathbb{R}^{B \times T \times C \times H \times W} \to \mathbb{R}^{B \times S \times D} \), where \( S \) is the number of encoded tokens, and \( D \) is the dimension of the latent representation. The encoder generates latent representations:
\[
f(R(v_i)) \in \mathbb{R}^{B \times S \times D} \quad \text{for the RGB video},
\]
\[
f'(\tilde{L}(v_i)) \in \mathbb{R}^{B \times S' \times D} \quad \text{for the masked RGB video},
\]
where \( S' \leq S \) due to the masking operation, and \( f' \) denotes the encoder operating on masked inputs.
\\
\\
\noindent\textbf{Primary Task of Deepfake Classification}
The latent representation \( f(R(v_i)) \) of the RGB video is passed to a classifier \( h : \mathbb{R}^{B \times S \times D} \to \mathbb{R}^{B \times 2} \) for binary deepfake classification. Each video \( v_i \) is associated with a ground-truth label \( y_i \in \{0, 1\} \), where \( y_i = 1 \) indicates a deepfake, and \( y_i = 0 \) indicates an authentic video. The classification loss is computed using cross-entropy:
{\footnotesize
\[
L_{\text{cls}} = -\frac{1}{B} \sum_{i=1}^B \left[ y_i \log(h(f(R(v_i)))_1) + (1 - y_i) \log(h(f(R(v_i)))_0) \right]
\]
}
where \( h(f(R(v_i)))_k \) denotes the predicted probability for class \( k \in \{0, 1\} \).
\\

\noindent\textbf{Auxiliary Task: Masked LDP2Video Prediction task}
The latent representation \( f'(\tilde{L}(v_i)) \) of the masked RGB video is passed to a shallow decoder \( g : \mathbb{R}^{B \times S' \times D} \to \mathbb{R}^{B \times T \times C \times H \times W} \), inspired by the VideoMAE approach\cite{videomae}. The decoder reconstructs the local texture patches from original RGB  \( L(v_i) \) masked video representation. Following the MAE framework, \( g \) takes \( f'(\tilde{L}(v_i)) \) and learnable masked tokens as input. Each token at the decoder’s output is linearly projected to a vector of pixel values representing a patch. The reconstructed video is:
\[
\hat{L}(v_i) = g(f'(\tilde{L}(v_i))).
\]
The reconstruction loss is the mean squared error (MSE) computed only over the masked patches:
{\footnotesize
\[
L_{\text{rec}} = \frac{1}{|\mathcal{M}_i|} \sum_{(t, h, w) \in \mathcal{M}_i} \left\| L(v_i)[:, t, :, h, w] - \hat{L}(v_i)[:, t, :, h, w] \right\|_2^2
\]
}
where \( \mathcal{M}_i \) is the set of masked patch indices in \( \tilde{L}(v_i) \), and \( |\mathcal{M}_i| \) is the number of masked patches.

The framework jointly optimizes the primary classification task and the auxiliary reconstruction task using a convex combination of losses:
\[
L = \lambda * L_{\text{cls}} + (1 - \lambda) * L_{\text{rec}}
\] 
where \( \lambda = 0.1 \), assigning a weight of 0.1 to the classification loss and 0.9 to the reconstruction loss, emphasizing the auxiliary task.

\begin{figure*}
    \centering
    \includegraphics[width=0.97\textwidth]{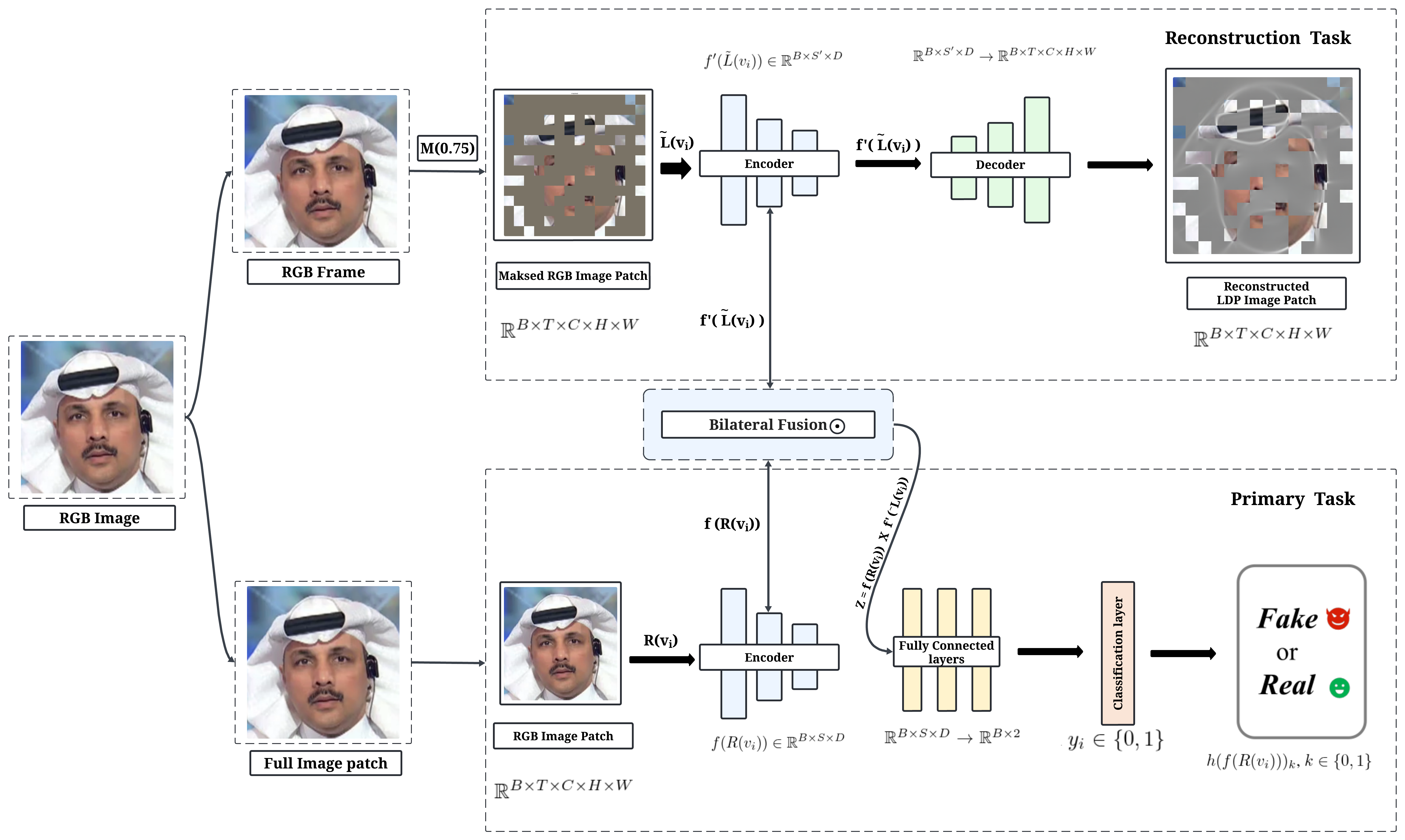} 
    \caption{Overview of the proposed Fusion-SSAT approach}
    \label{fig:ee}
\end{figure*}

\noindent \textbf{Fusion strategy:}  A joint feature embedding strategy is introduced, where features from both the LDP\cite{LDP} and RGB denoted as \( f'(\tilde{L}(v_i)) \) and \( f(R(v_j)) \), respectively are combined via element-wise multiplication: \( z = f'(\tilde{L}(v_i)) \odot f(R(v_j)) \). The fused feature vector \( z \) is then fed into the classifier \( h \). This joint embedding enables the classifier to concurrently leverage fine-grained texture information from LDP features\cite{LDP} and high-level semantic cues from RGB inputs, thereby improving classification performance compared to the earlier independent processing approach as shown in Figure~\ref{fig:ee} This formulation ensures that the shared ViT encoder\cite{dosovitskiy2021imageworth16x16words} learns robust representations for deepfake detection from RGB videos while leveraging self-supervised learning to reconstruct masked patches in LDP videos, enhancing generalization and capturing texture-based features critical for deepfake analysis.
\section{Experimental results}
\vspace{1mm}

In this section, we described the experiments' details and results on the proposed model Fusion-SSAT in comparison with state-of-the-art deepfake detectors. Section 4.1 describes the details of datasets on which proposed models has been evaluated, While Section 4.2 describes the implementation details, hardware and evaluation metrics used in reporting the performance of proposed models along with state-of-the-art models. Section 4.3 describes results comparison between proposed model with existing state-of-the-art models. 
\subsection{Datasets}
We evaluted our proposed model Fusion-SSAT on FaceForensics++\cite{ff++}, Celeb-DF v1\cite{Celeb_DF}, Celeb-DF v2\cite{Celeb_DF}, FaceShifter\cite{Faceshifter}, deep fake detection\cite{ff++}, UADFV\cite{UADFV}, DF40\cite{df40} datasets. All proposed models are trained on FF++ c23, and tested on cross-domain datasets mentioned. Detailed description of each dataset is given below: 
\\
\textbf{FaceForensics++\cite{ff++}:} is a huge benchmark dataset that consists of 1000 real videos and 4000 fake videos, while 720 is used as the train split, 140 each for test and validation splits. Dataset contains manipulations created with state-of-the-art methods, namely, Face2Face, FaceSwap, DeepFakes, and Neuraltextures. There exist three variants of FF++ relating to video compression levels i.e raw, lightly compressed (c23), and highly compressed (c40).

\noindent\textbf{Celeb-DF v1, v2\cite{Celeb_DF}:} includes high quality face-swapping videos, while Celeb-DF v1 has 408 original videos and 795 fake videos while v2 has 590 original videos and 5639 fake videos. 

\noindent\textbf{Faceshifter\cite{Faceshifter}:} is a method to created high fidelity and occlusion aware face swapping videos. It is a subset of the FaceForensics++ dataset, consists of 1000 fake videos. 

\noindent\textbf{Deep Fake Detection\cite{ff++}:} is a dataset which is developed by google, includes 363 real videos and 3000 fake videos. Now it's available as part of FaceForensics++ dataset. \textbf{UADFV\cite{UADFV}:} dataset consists of 98 videos, in which 49 videos are real and 49 videos are fake. 

\noindent\textbf{DF40\cite{df40}:} is a huge dataset which compress different state-of-the-art deepfake generation techniques Contains 0.1M+ fake videos. Used real videos from FaceForensics++ and Celeb-DF dataset's for creating fake videos. Dataset is broadly categorized into 4 parts namely, Face-Swapping (FS), Face-Reenactment (FR), Entire Face Synthesis (EFS), Face Editing (FE). DF40 is created by 40 deepfake techniques, including 10 FS methods, 13 FR methods, 12 EFS methods, and 5 FE methods. 
\subsection{Implementation details and Evaluation Metrics}
All proposed models are trained on images/videos of dimension 224 x 224 pixels using patches of 16 x 16 pixels as input (LDP or RGB) to the respective task (classification or reconstruction) of the model. The training is conducted in batch size of 8 on an NVIDIA A100 GPU utilizing the stochastic gradient descent optimizer for weight adjustments. The learning rate is determined by a cosine schedule that starts at 0.00005 and targets a minimum learning rate of 1e-6. The learning rate is consistently diminished by incorporating a decay rate of 0.05 across each of the 75 training epochs the model experiences. The stochastic depth is implemented with an initial drop path rate of 0.01, and the different loss terms are balanced using a Lambda (\( \lambda \)) value of 0.1. A masking ratio of 0.75 has been established. The official partitions are utilized for all datasets.
\subsection{Results and discussion}
We have evaluated our proposed model Fusion-SSAT in two different scenario's. In section 4.3.1 all models are trained on FaceForensics++ c23\cite{ff++}, and tested on in-domain and cross-domain datasets which includes Celeb-DF v1\cite{Celeb_DF}, Celeb-DF v2\cite{Celeb_DF}, DFD\cite{ff++}, FaceShifter\cite{Faceshifter}, UADFV\cite{UADFV}. In section 4.3.2 all model's are trained on DF40 train set of FF++ domain with FS (FF), FR(FF), EFS(FF) forgery data and evaluated on within-forgery, cross-forgery (FF)\cite{ff++} and cross-domain(CDF)\cite{Celeb_DF} of all forgery methods. In our proposed model, Fusion-SSAT naming convention, the modality preceding the hyphen signifies the masked video input supplied to the reconstruction task, whereas the modality following the hyphen indicates the domain in which the prediction occurs. For instance, In RGB-LDP Fusion-SSAT, the masked video input originates from the RGB modality, while the model predicts in the LDP. We employed the AUC as the evaluation metric. 
The best AUC scores achieved by our models are \underline {underlined}, while the best SOTA results are \textbf{\textit{italicized}}, and the highest-performing method in each column is indicated in \textbf{bold}.

\begin{table*}
\caption{\em \small In-domain and cross-domain evaluations using the AUC metric.
All detectors are trained on FF-c23 and evaluated on other data. 
``Avg." donates the average AUC for within-domain and cross-domain evaluation, and the overall results.     
}
\renewcommand\arraystretch{1.8}
\setlength\tabcolsep{2.5pt}
\centering
\vspace{2mm}
\resizebox{\linewidth}{!}{
\begin{tabular}{>{\centering\arraybackslash}m{3.7cm}>{\centering\arraybackslash}m{2.8cm}*{6}{>{\centering\arraybackslash}p{1.2cm}}*{1}{>{\centering\arraybackslash}p{1.0cm}}*{5}{>{\centering\arraybackslash}p{1.2cm}}>{\centering\arraybackslash}p{1.0cm}}
\toprule
\multirow{2}{*}{\centering\textbf{Detector}} & \multirow{2}{*}{\centering\textbf{Backbone}} & \multicolumn{7}{c}{\textbf{ In-Domain Evaluation}} & \multicolumn{6}{c}{\textbf{Cross Domain Evaluation}}\\
\cmidrule(lr){3-9} \cmidrule(lr){10-15}
&& FF-c23 & FF-c40 & FF-DF & FF-F2F & FF-FS & FF-NT & Avg. & CDFv1 & CDFv2 & DFD & Fsh & UADFV & Avg. \\
\midrule
Meso4~\cite{afchar2018mesonet}'WIFS-18 & MesoNet\cite{afchar2018mesonet} & 0.6077 & 0.5920 & 0.6771 & 0.6170 & 0.5946 & 0.5701 & 0.6097 & 0.7358 & 0.6091 & 0.5481 & 0.5660 & 0.7150 & 0.6348 \\
MesoIncep~\cite{afchar2018mesonet}'WIFS-18 & MesoNet\cite{afchar2018mesonet} & 0.7583 & 0.7278 & 0.8542 & 0.8087 & 0.7421 & 0.6517 & 0.7571 & 0.7366 & 0.6966 & 0.6069 & 0.6438 & 0.9049 &0.7177 \\
CNN-Aug~\cite{wang2020cnn}'CVPR-20 & ResNet\cite{he2016deep} & 0.8493 & 0.7846 & 0.9048 & 0.8788 & 0.9026 & 0.7313 & 0.8419 & 0.7420 & 0.7027 & 0.6464 & 0.5985 & 0.8739 & 0.7127  \\
Xception~\cite{ff++}'ICML-19 & Xception\cite{chollet2017xceptiondeeplearningdepthwise} & 0.9637 & 0.8261 & 0.9799 & 0.9785 & 0.9833 & 0.9385 & 0.9450 & 0.7794 & 0.7365 & 0.8163 & 0.6249 & 0.9379 & 0.7790 \\
EfficientB4~\cite{tan2020efficientnetrethinkingmodelscaling}'ICCV-19 &Efficient\cite{tan2020efficientnetrethinkingmodelscaling} & 0.9567 & 0.8150 & 0.9757 & 0.9758 & 0.9797 & 0.9308 & 0.9389 & 0.7909 & 0.7487 & 0.8148 & 0.6162 & 0.9472 & 0.7835\\
Capsule\cite{capsule}'CASSP-19 & Capsule\cite{capsule} & 0.8421 & 0.7040 & 0.8669 & 0.8634 & 0.8734 & 0.7804 & 0.8217 & 0.7909 & 0.7472 & 0.6841 & 0.6465 & 0.9078 & 0.7553 \\
FWA\cite{DSP-FWA}'CVPRW-19 & Xception\cite{chollet2017xceptiondeeplearningdepthwise} & 0.8765 & 0.7357 & 0.9210 & 0.9000 & 0.8843 & 0.8120 & 0.8549 & 0.7897 & 0.6680 & 0.7403 & 0.5551 & 0.8539 & 0.7214 \\
X-ray\cite{Face-X-Ray}'CVPR-20 & HRNet\cite{HRNet} & 0.9592 & 0.7925 & 0.9794 & 0.9872 & 0.9871 & 0.9290 & 0.9391 & 0.7093 & 0.6786 & 0.7655 & 0.6553 & 0.8989 & 0.7415 \\
FFD\cite{FFD}'CVPR-20 & Xception\cite{chollet2017xceptiondeeplearningdepthwise} & 0.9624 & 0.8237 & 0.9803 & 0.9784 & 0.9853 & 0.9306 & 0.9434 & 0.7840 & 0.7435 & 0.8024 & 0.6056 & 0.9450 & 0.7761 \\
CORE\cite{Ni_2022_CVPR}'CVPRW-20 & Xception\cite{chollet2017xceptiondeeplearningdepthwise} & 0.9638 & 0.8194 & 0.9787 & 0.9803 & 0.9823 & 0.9339 & 0.9431 & 0.7798 & 0.7428 & 0.8018 & 0.6032 & 0.9412 & 0.7738 \\
Recce\cite{Cao_2022_CVPR}'CVPR-20 & Designed & 0.9621 & 0.8190 & 0.9797 & 0.9779 & 0.9785 & 0.9357 & 0.9422 & 0.7677 & 0.7319 & 0.8119 & 0.6095 & 0.9446 & 0.7731 \\
F3Net~\cite{qian2020thinking}'ECCV-20 & Xception\cite{chollet2017xceptiondeeplearningdepthwise} & 0.9635 & 0.8271 & 0.9793 & 0.9796 & 0.9844 & 0.9354 & 0.9449 & 0.7769 & 0.7352 & 0.7975 & 0.5914 & 0.9347 & 0.7614 \\
SPSL~\cite{SPSL}'CVPR-21 & Xception\cite{chollet2017xceptiondeeplearningdepthwise} & 0.9610 & 0.8174 & 0.9781 & 0.9754 & 0.9829 & 0.9299 & 0.9408 & 0.8150 & 0.7650 & 0.8122 & 0.6437 & 0.9424 & \textbf{\textit{0.7956}} \\
SRM~\cite{SRM}'CVPR-21 & Xception\cite{chollet2017xceptiondeeplearningdepthwise} & 0.9576 & 0.8114 & 0.9733 & 0.9696 & 0.9740 & 0.9295 & 0.9359 & 0.7926 & 0.7552 & 0.8120 & 0.6014 & 0.9427 & 0.6222 \\
UCF\cite{ucf}'ICCV-23 & Xception\cite{chollet2017xceptiondeeplearningdepthwise} & 0.9705 & 0.8399 & 0.9883 & 0.9840 & 0.9896 & 0.9441 & \textbf{\textit{0.9527}} & 0.7793 & 0.7527 & 0.8074 & 0.6462 & 0.9528 & \textbf{\textit{0.7877}} \\
MARLIN~\cite{MARLIN}'CVPR-23 & ViT-L & 0.9370 & 0.8059 & 0.9399 & 0.9250 & 0.9580 & 0.9708 & 0.9228 & 0.7140 & 0.7030 & 0.6850 & 0.6671 & 0.8150 & 0.7168 \\
AVAD~\cite{AVAD}'CVPR-23 & Designed & 0.9410 & 0.7980 & 0.9872 & 0.9534 & 0.9267 & 0.9093 & 0.9193 & 0.7382 & 0.7089 & 0.6892 & 0.7120 & 0.8910 & 0.7479 \\
DF-Adapter~\cite{deepfakeadapter}'IJCV-24 & ViT/ViT-Adapter\cite{dosovitskiy2021imageworth16x16words} & 0.9862 & 0.9683 & 0.9984 & 0.9976 & 0.9933 & 0.9597 & \textbf{\textit{0.9839}} & 0.7174 & 0.7071 & 0.7194 & 0.7292 & 0.8948 & 0.7536 \\
StyleGRU~\cite{StyleGRU}'CVPR-24 & StyleGRU\cite{StyleGRU} & 0.8297 & 0.7091 & 0.9400 & 0.6850 & 0.8880 & 0.8060 & 0.8297 & 0.6250 & 0.6051 & 0.7220 & 0.9070 & 0.8145 & 0.7347 \\
Fairness~\cite{fairness}'CVPR-24 & Xception\cite{chollet2017xceptiondeeplearningdepthwise} & 0.9828 & 0.8091 & 0.9905 & 0.9865 & 0.9923 & 0.9635 & 0.9541 & 0.7442 & 0.7002 & 0.8482 & 0.8342 & 0.8523 & 0.7958 \\
OPR~\cite{OPR}'NeurIPS-24 & EfficientNetB4\cite{tan2020efficientnetrethinkingmodelscaling} & 0.9591 & 0.6078 & 0.9861 & 0.9272 & 0.9272 & 0.9071 & 0.8858 & 0.9094 & 0.8448 & 0.8091 & 0.7190 & 0.8581 & 0.8281 \\
PFF~\cite{PFF}'ICLR-24 & Xception\cite{chollet2017xceptiondeeplearningdepthwise} & 0.8518 & 0.6982 & 0.8591 & 0.9281 & 0.8271 & 0.8561 & 0.8367 & 0.7721 & 0.7041 & 0.7826 & 0.6982 & 0.8125 & 0.7539 \\
SSAT~\cite{das-limiteddatavit-wacv2024}'WACV-24 & VideoMAE/ViT-B &  0.9682 & 0.7438 & 0.9991 & 0.9603 & 0.9922 & 0.9215 & 0.9308 & 0.8073 & 0.7757 & 0.8109 & 0.9902 & 0.8920 & 0.8552 \\
LDP-RGBFNet\cite{ICPR}'ICPRW-24 & VideoMAE/ViT-B & 0.6766 & 0.6566 & 0.8622 & 0.6047 & 0.6112 & 0.6281 & 0.6732 & 0.4607 & 0.6828 & 0.5654 & 0.8052 & 0.7344 & 0.6497 \\
LDP-RGBFNet\cite{ICPR}'ICPRW-24 & VideoMAE/ViT-L & 0.9868 & 0.7687 & 0.9998 & 0.9871 & 0.9972 & 0.9631 & 0.9505 & 0.8256 & 0.8165 & 0.8973 & 0.9933 & 0.9545 & 0.8974 \\
LDP-LDPFNet\cite{ICPR}'ICPRW-24 & VideoMAE/ViT-B& 0.9687 & 0.7727 & 0.9995 & 0.9667 & 0.9893 & 0.9193 & 0.9360 & 0.7676 & 0.7499 & 0.7537 & 0.9915 & 0.8971 & 0.8319 \\
LDP-LDPFNet\cite{ICPR}'ICPRW-24 & VideoMAE/ViT-L& 0.9857 & 0.7673 & 0.9996 & 0.9844 & 0.9969 & 0.9617 & 0.9493 & 0.9193 & 0.8676 & 0.8766 & 0.9946 & 0.9517 & 0.9219 \\
RGB-LDPFNet\cite{ICPR}'ICPRW-24 & VideoMAE/ViT-B & 0.9740 & 0.7724 & 0.9996 & 0.9688 & 0.9914 & 0.9364 & 0.9404 & 0.8055 & 0.7550 & 0.7634 & 0.9904 & 0.9077 & 0.8444 \\
RGB-LDPFNet\cite{ICPR}'ICPRW-24 & VideoMAE/ViT-L & 0.9815 & 0.8283 & 0.9996 & 0.9828 & 0.9940 & 0.9496 & 0.9559 & 0.8727 & 0.8477 & 0.7955 & 0.9902 & 0.9409 & 0.8894 \\
\midrule
\rowcolor{gray!20}
RGB-LDP Fusion-SSAT & VideoMAE/ViT-B & 0.9491 & 0.7796 & 0.9941 & 0.9451 & 0.9866 & 0.8710 & 0.9209 & 0.7170 & 0.7659 & 0.7551 & 0.9807 & 0.9311 & 0.8299 \\

\rowcolor{gray!20}
RGB-LDP Fusion-SSAT & VideoMAE/ViT-L & 0.9865 & \textbf{{0.8349}} & 0.9996 & \textbf{{0.9921}} & \textbf{{0.9978}} & 0.9571 & \underline{0.9613} & 0.8740 & 0.8256 & 0.8122 & \textbf{{0.9962}} & \textbf{{0.9570}} & \underline{0.8930} \\
\bottomrule
\end{tabular}
}
\label{tab:deepfakebench}
\vspace{2mm}
\end{table*}

\begin{table*}
\tiny
\centering
\centering
\scriptsize
\setlength{\tabcolsep}{9.8pt} 
\renewcommand{\arraystretch}{1.2} 
\caption{\em \small \textbf{Cross-forgery and Cross-domain evaluation:} Models trained on DF40 FF domain with different forgery methods (FS, FR, EFS). Evaluation is performed on both FF and CDF domains.}
\label{tab:merged}
\begin{tabular}{c|c|cccc|cccc}
\hline
\multirow{2}{*}{\textbf{Training Set}} & \multirow{2}{*}{\textbf{Model}} & \multicolumn{4}{c|}{\textbf{Testing Set (FF)}} & \multicolumn{4}{c}{\textbf{Testing Set (CDF)}} \\ \cline{3-10}
                              &                        & FS & FR & EFS & Avg. (FF) & FS & FR & EFS & Avg. (CDF) \\ \hline
\multirow{6}{*}{\textbf{FS (FF)}}     
& Xception\cite{chollet2017xceptiondeeplearningdepthwise}'CVPR-17 & 0.991 & 0.892 & 0.810 & 0.898 & 0.922 & 0.657 & 0.642 & \textbf{\textit{0.740}} \\
& CLIP\cite{clip}'ICML-21 & 0.996 & 0.908 & 0.837 & \textbf{\textit{0.914}} & 0.967 & 0.744 & 0.730 & 0.814 \\
& SRM\cite{SRM}'CVPR-21 & 0.988 & 0.867 & 0.703 & 0.853 & 0.919 & 0.621 & 0.603 & 0.714 \\
& SPSL\cite{SPSL}'CVPR-21 & 0.987 & 0.849 & 0.735 & 0.857 & 0.938 & 0.656 & 0.648 & \textbf{\textit{0.747}} \\
& RECCE\cite{Cao_2022_CVPR}'CVPR-20 & 0.991 & 0.855 & 0.758 & 0.868 & 0.926 & 0.632 & 0.610 & 0.723 \\
& RFM\cite{RFM}'CVPR-21 & 0.992 & 0.884 & 0.821 & \textbf{\textit{0.899}} & 0.939 & 0.637 & 0.628 & 0.735 \\
& SSAT~\cite{das-limiteddatavit-wacv2024}'WACV-24  & 0.988 & 0.963 & 0.972 & 0.974 & 0.985 & 0.860 & 0.990 & 0.945 \\
& LDP-RGBFNet ViT-B\cite{ICPR}'ICPRW-24  & 0.458 & 0.354 & 0.322 & 0.378 & 0.577 & 0.432 & 0.711 & 0.573 \\
& LDP-RGBFNet ViT-L\cite{ICPR}'ICPRW-24 & 0.991  & 0.950 & 0.954 & 0.965 & 0.994 & 0.882 & 0.992 & 0.956 \\
& LDP-LDPFNet ViT-B\cite{ICPR}'ICPRW-24 & 0.991 & 0.978 & 0.899 & 0.956 & 0.990 & 0.895 & 0.968 & 0.951 \\
& LDP-LDPFNet ViT-L\cite{ICPR}'ICPRW-24 & 0.993  & 0.970 & 0.932 & 0.965 & 0.995 & 0.916 & 0.988 & 0.966 \\
& RGB-LDPFNet ViT-B\cite{ICPR}'ICPRW-24 & 0.994 & 0.980 & 0.870 & 0.948 & 0.992 & 0.913 & 0.976 & 0.960 \\
& RGB-LDPFNet ViT-L\cite{ICPR}'ICPRW-24 & 0.994 & 0.981 & 0.905 & 0.960 & 0.998 & 0.951 & 0.979 & \underline{0.976} \\

& \cellcolor{gray!20}RGB-LDP Fusion-SSAT ViT-B & \cellcolor{gray!20}0.978 & \cellcolor{gray!20}0.982 & \cellcolor{gray!20}0.950 & \cellcolor{gray!20}\underline{0.970} & \cellcolor{gray!20}0.995 & \cellcolor{gray!20}0.949 & \cellcolor{gray!20}0.989 & \cellcolor{gray!20}0.977 \\

& \cellcolor{gray!20}RGB-LDP Fusion-SSAT ViT-L & \cellcolor{gray!20}\textbf{{0.996}} & \cellcolor{gray!20}\textbf{{0.993}} & \cellcolor{gray!20}0.920 & \cellcolor{gray!20}\underline{0.970} & \cellcolor{gray!20}\textbf{{0.998}} & \cellcolor{gray!20}\textbf{{0.976}} & \cellcolor{gray!20}0.984 & \cellcolor{gray!20}\underline{0.986} \\ \hline

\multirow{10}{*}{\textbf{FR (FF)}}     
& Xception\cite{chollet2017xceptiondeeplearningdepthwise}'CVPR-17                & 0.838 & 0.996 & 0.670 & 0.835 & 0.481 & 0.857 & 0.369 & 0.569 \\
& CLIP\cite{clip}'ICML-21                   & 0.932 & 0.999 & 0.798 & \textbf{\textit{0.910}} & 0.638 & 0.933 & 0.209 & \textbf{\textit{0.593}} \\
& SRM\cite{SRM}'CVPR-21                    & 0.893 & 0.998 & 0.698 & 0.863 & 0.454 & 0.869 & 0.326 & 0.550 \\
& SPSL\cite{SPSL}'CVPR-21                   & 0.901 & 0.998 & 0.695 & 0.865 & 0.479 & 0.852 & 0.256 & 0.529 \\
& RECCE\cite{Cao_2022_CVPR}'CVPR-20                  & 0.865 & 0.997 & 0.716 & 0.859 & 0.452 & 0.881 & 0.332 & 0.555 \\
& RFM\cite{RFM}'CVPR-21                    & 0.892 & 0.999 & 0.776 & \textbf{\textit{0.889}} & 0.492 & 0.882 & 0.359 & \textbf{\textit{0.578}} \\

& SSAT~\cite{das-limiteddatavit-wacv2024}'WACV-24  & 0.991 & 0.997 & 0.937 & 0.975 & 0.968 & 0.982 & 0.955 & 0.968 \\
& LDP-RGBFNet ViT-B\cite{ICPR}'ICPRW-24  & 0.379 & 0.661 & 0.017 & 0.352 & 0.348 & 0.550 & 0.096 & 0.331 \\
& LDP-RGBFNet ViT-L\cite{ICPR}'ICPRW-24 & \textbf{{0.996}} & \textbf{{0.996}} & 0.955 & \underline{0.982} & \textbf{{0.994}} & 0.994 & 0.980 & \underline{0.989} \\

& LDP-LDPFNet ViT-B\cite{ICPR}'ICPRW-24 & 0.980 & 0.993 & 0.919 & 0.964 & 0.970 & 0.982 & 0.932 & 0.961 \\

& LDP-LDPFNet ViT-L\cite{ICPR}'ICPRW-24 & 0.988 & 0.993 & 0.924 & \underline{0.968} & 0.991 & 0.992 & 0.966 & \underline{0.983} \\

& RGB-LDPFNet ViT-B\cite{ICPR}'ICPRW-24 & 0.972 & 0.987 & 0.914 & 0.958 & 0.972 & 0.979 & 0.962 & 0.971 \\

& RGB-LDPFNet ViT-L\cite{ICPR}'ICPRW-24 & 0.975 & 0.985 & 0.894 & 0.951 & 0.981 & 0.980 & 0.949 & 0.970 \\

& \cellcolor{gray!20}RGB-LDP Fusion-SSAT ViT-B& \cellcolor{gray!20}0.936 & \cellcolor{gray!20}0.954 & \cellcolor{gray!20}0.894 & \cellcolor{gray!20}0.928 & \cellcolor{gray!20}0.964 & \cellcolor{gray!20}0.973 & \cellcolor{gray!20}0.947 & \cellcolor{gray!20}0.961 \\

& \cellcolor{gray!20}RGB-LDP Fusion-SSAT ViT-L& \cellcolor{gray!20}0.984 & \cellcolor{gray!20}0.994 & \cellcolor{gray!20}0.890 & \cellcolor{gray!20}0.956 & \cellcolor{gray!20}0.990 & \cellcolor{gray!20}\textbf{{0.995}} & \cellcolor{gray!20}0.959 & \cellcolor{gray!20}0.981 \\ \hline

\multirow{10}{*}{\textbf{EFS (FF)}}    
& Xception\cite{chollet2017xceptiondeeplearningdepthwise}'CVPR-17                & 0.665 & 0.807 & 0.999 & 0.824 & 0.586 & 0.594 & 0.983 & 0.721 \\
& CLIP\cite{clip}'ICML-21                   & 0.688 & 0.889 & 0.999 & \textbf{\textit{0.859}} & 0.617 & 0.735 & 0.988 & \textbf{\textit{0.780}} \\
& SRM\cite{SRM}'CVPR-21                    & 0.596 & 0.776 & 0.999 & 0.790 & 0.589 & 0.620 & 0.964 & 0.724 \\
& SPSL\cite{SPSL}'CVPR-21                   & 0.659 & 0.811 & 0.999 & 0.823 & 0.635 & 0.651 & 0.975 & 0.754 \\
& RECCE\cite{Cao_2022_CVPR}'CVPR-20                 & 0.691 & 0.801 & 0.999 & \textbf{\textit{0.830}} & 0.623 & 0.603 & 0.984 & 0.737 \\
& RFM\cite{RFM}'CVPR-21                    & 0.653 & 0.795 & 0.999 & 0.816 & 0.644 & 0.666 & 0.981 & \textbf{\textit{0.764}} \\

& SSAT~\cite{das-limiteddatavit-wacv2024}'WACV-24  & 0.997 & 0.997 & 1.0 & 0.998 & 0.995 & 0.959 & 1.0 & 0.985 \\
& LDP-RGBFNet ViT-B\cite{ICPR}'ICPRW-24  & 0.924 & 0.837 & 1.0 & 0.920 & 0.823 & 0.719 & 1.0 & 0.847 \\
& LDP-RGBFNet ViT-L\cite{ICPR}'ICPRW-24 & 0.999 & 0.998 & 1.0 & 0.999 & 0.997 & 0.981 & 1.0 & 0.993 \\
& LDP-LDPFNet ViT-B\cite{ICPR}'ICPRW-24 & 0.999 & 0.997 & 1.0 & 0.998 & 0.997 & 0.979 & 1.0 & \underline{0.992} \\
& LDP-LDPFNet ViT-L\cite{ICPR}'ICPRW-24 & 0.999 & 0.998 & 1.0 & 0.999 & 0.997 & 0.981 & \textbf{{1.0}} & 0.993 \\
& RGB-LDPFNet ViT-B\cite{ICPR}'ICPRW-24 & 0.999 & 0.996 & 0.999 & \underline{0.998} & 0.996 & 0.965 & 0.990 & 0.983 \\
& RGB-LDPFNet ViT-L\cite{ICPR}'ICPRW-24 & 0.999 & 0.998 & 0.999 & \underline{0.998} & 0.995 & 0.980 & 0.990 & 0.988 \\

& \cellcolor{gray!20}RGB-LDP Fusion-SSAT ViT-B & \cellcolor{gray!20}0.996 & \cellcolor{gray!20}0.964 & \cellcolor{gray!20}0.999 & \cellcolor{gray!20}0.986 & \cellcolor{gray!20}0.971 & \cellcolor{gray!20}0.840 & \cellcolor{gray!20}0.990 & \cellcolor{gray!20}0.933 \\

& \cellcolor{gray!20}RGB-LDP Fusion-SSAT ViT-L & \cellcolor{gray!20}\textbf{{0.999}} & \cellcolor{gray!20}\textbf{{0.999}} & \cellcolor{gray!20}0.999 & \cellcolor{gray!20}\underline{0.999} & \cellcolor{gray!20}\textbf{{0.997}} & \cellcolor{gray!20}\textbf{{0.986}} & \cellcolor{gray!20}0.990 & \cellcolor{gray!20}\underline{0.991} \\ \hline

\multirow{3}{*}{All}    
& MARALIN \cite{MARLIN}'CVPR-23 & - & - & - & 0.981 & - & - & - & 0.796 \\
& GrDT \cite{GrDT}'WACVW-25 & - & - & - & 0.986 & - & - & - & 0.824 \\
& \cellcolor{gray!20} Ours & \cellcolor{gray!20} - & \cellcolor{gray!20} - & \cellcolor{gray!20} - & \cellcolor{gray!20}0.999 & \cellcolor{gray!20} - & \cellcolor{gray!20} - & \cellcolor{gray!20} - & \cellcolor{gray!20} 0.991 \\ \hline
\end{tabular}
\end{table*}

\subsubsection{Experimental results on Within-Domain and Cross-Domain}
Table~\ref{tab:deepfakebench} exhibits the experimental results of our proposed model trained on FF++ c23\cite{ff++} (Deepfakes(FF-DF), Face2Face(FF-F2F), FaceSwap(FF-FS), Neuraltextures(FF-NT)) and evaluated on within domain evaluation and cross domain datasets. We benchmark our results against deepfakebench's\cite{DeepfakeBench} spatial detector's which exhibited better performance than existing state-of-the-art deepfake detectors.

Column-1 defines the specifications of the detector, whereas Column-2 defines the backbone network employed in the respective detector. Column-3, 4 defines the experimental results (AUC score) of with domain evaluation and cross-domain evaluation, that includes test scores on the complete datasets FF-c23\cite{ff++}, FF-c40\cite{ff++}, CDFv1\cite{Celeb_DF}, CDFv2\cite{Celeb_DF}, DFD\cite{ff++}, Fsh\cite{Faceshifter}, UADFV\cite{UADFV} and specific forging methods. The 'Avg' column outlines the performance of each detector over all datasets, both within-domain and cross-domain.     

\begin{figure}
    \centering
    \begin{subfigure}{0.4\textwidth}
        \centering
        \includegraphics[width=1\linewidth]{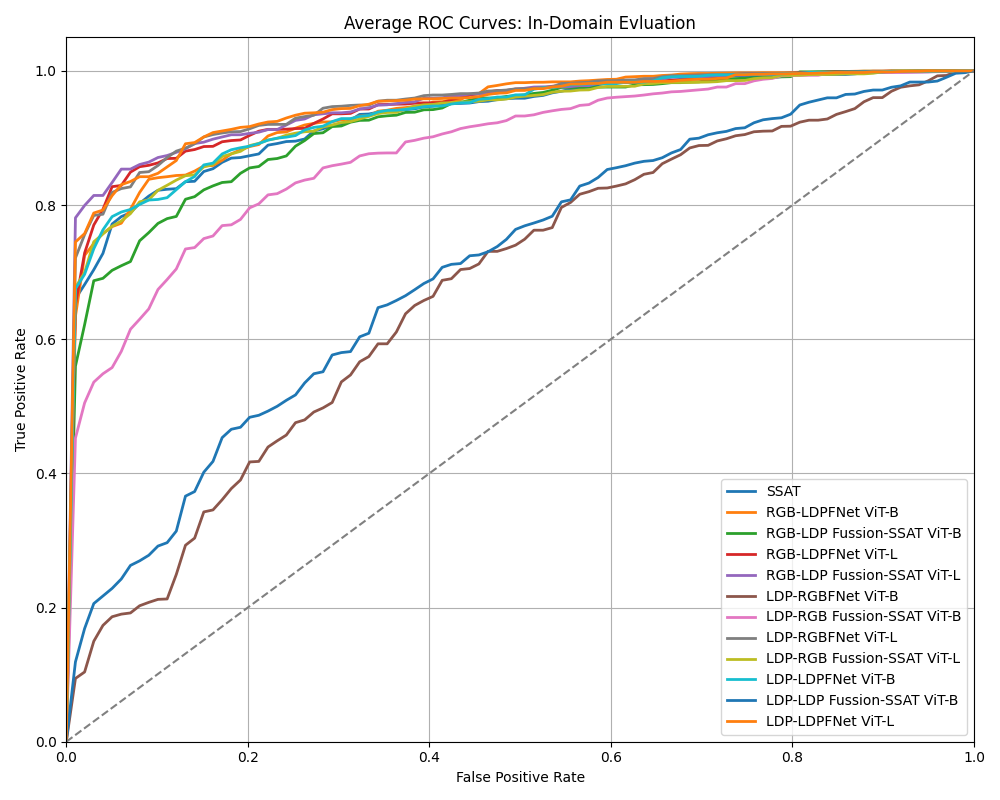}
        \caption{In-domain evaluation on FF++ dataset}
        \label{fig:roc-withindomain}
    \end{subfigure}\hfill
    \begin{subfigure}{0.4\textwidth}
        \centering
        \includegraphics[width=1\linewidth]{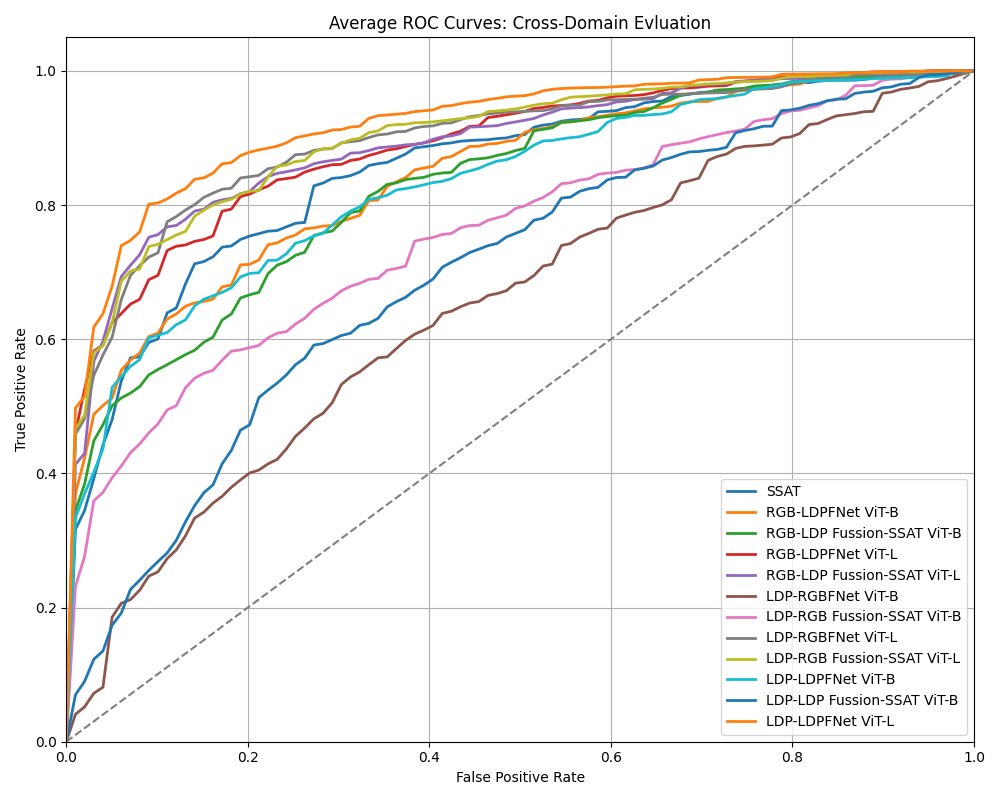}
        \caption{Cross-domain evaluation on CDF, DFD, UADFV datasets}
        \label{fig:roc-crossdomain}
    \end{subfigure}
    \caption{ROCs for in-domain and cross-domain evaluations.}
    \label{fig:roc-evaluations}
\end{figure}

\begin{figure}
    \centering
    \begin{subfigure}{0.4\textwidth}
        \centering
        \includegraphics[width=1\linewidth]{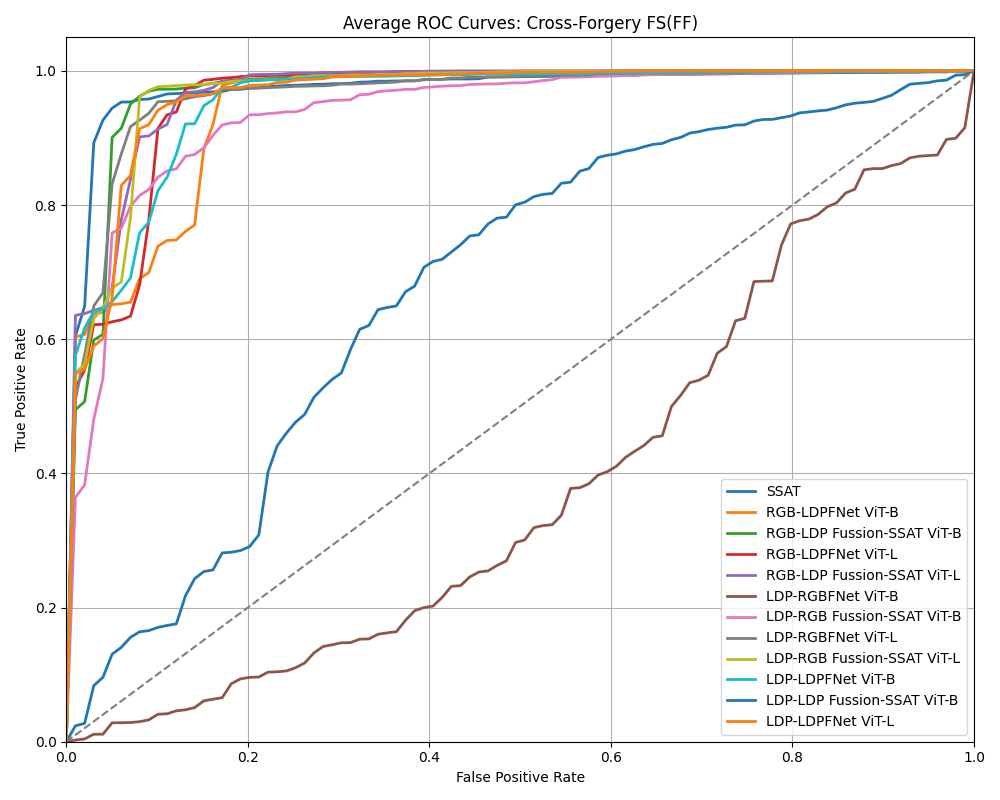}
        \caption{FS (FF)}
        \label{fig:cf1}
    \end{subfigure}\hfill
    \begin{subfigure}{0.4\textwidth}
        \centering
        \includegraphics[width=1\linewidth]{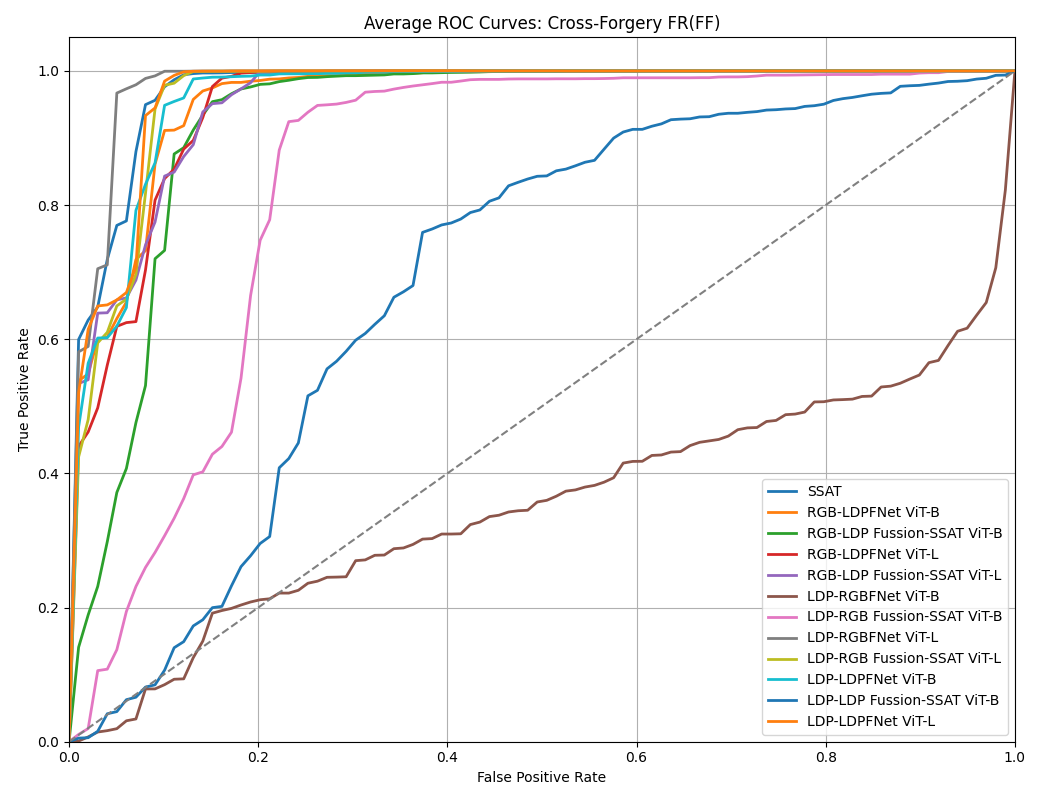}
        \caption{FR (FF)}
        \label{fig:cf2}
    \end{subfigure}\hfill
    \begin{subfigure}{0.4\textwidth}
        \centering
        \includegraphics[width=1\linewidth]{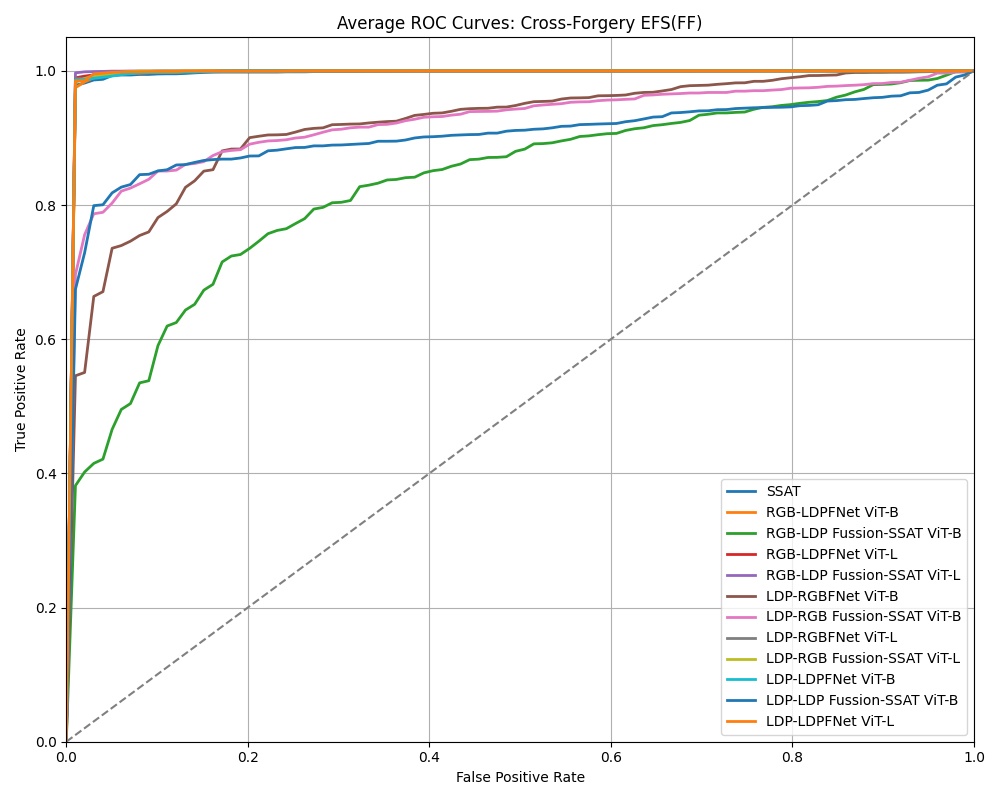}
        \caption{EFS (FF)}
        \label{fig:cf3}
    \end{subfigure}
    \caption{ROCs of cross-forgery evaluations on DF40 dataset.}
    \label{fig:cross_forgery_roc-DF}
\end{figure}

\begin{figure}[p]
    \centering
    \begin{subfigure}{0.4\textwidth}
        \centering
        \includegraphics[width=\linewidth]{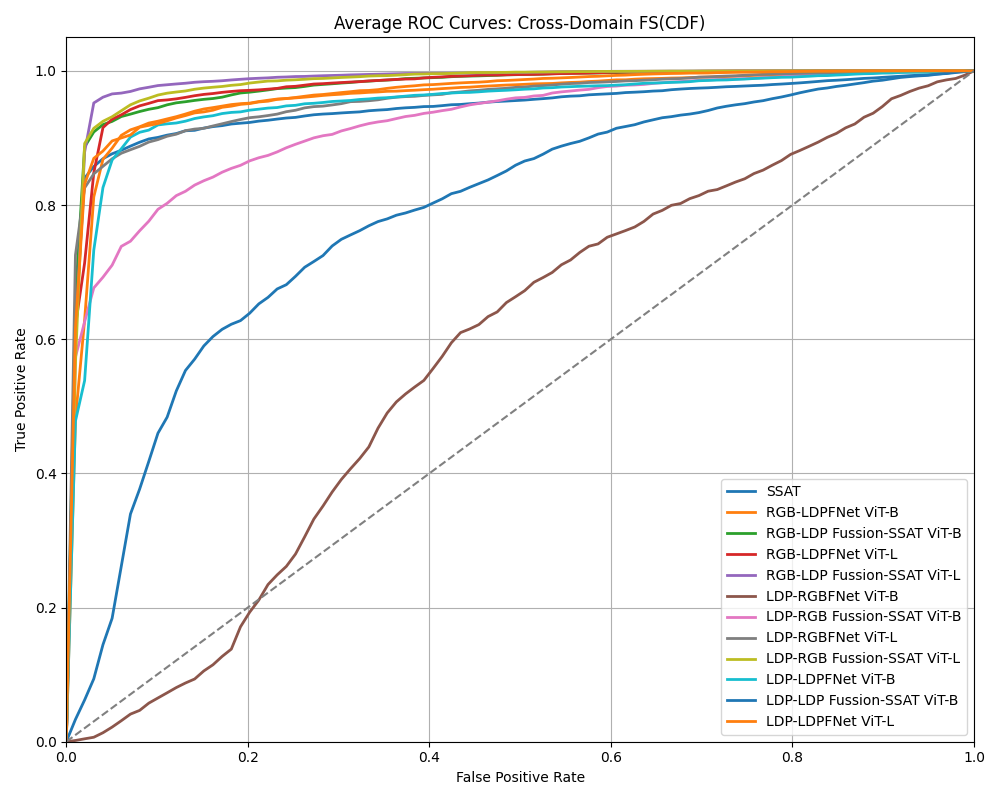}
        \caption{FS (CDF)}
        \label{fig:cd1}
    \end{subfigure}\hfill
    \begin{subfigure}{0.4\textwidth}
        \centering
        \includegraphics[width=\linewidth]{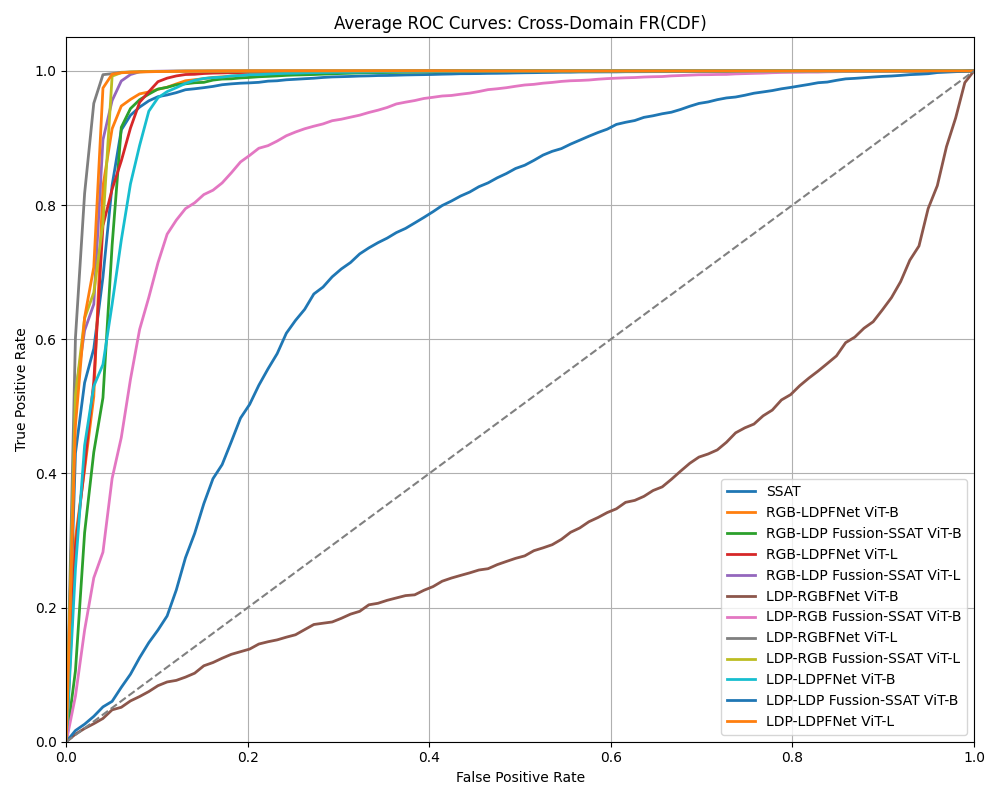}
        \caption{FR (CDF)}
        \label{fig:cd2}
    \end{subfigure}\hfill
    \begin{subfigure}{0.4\textwidth}
        \centering
        \includegraphics[width=\linewidth]{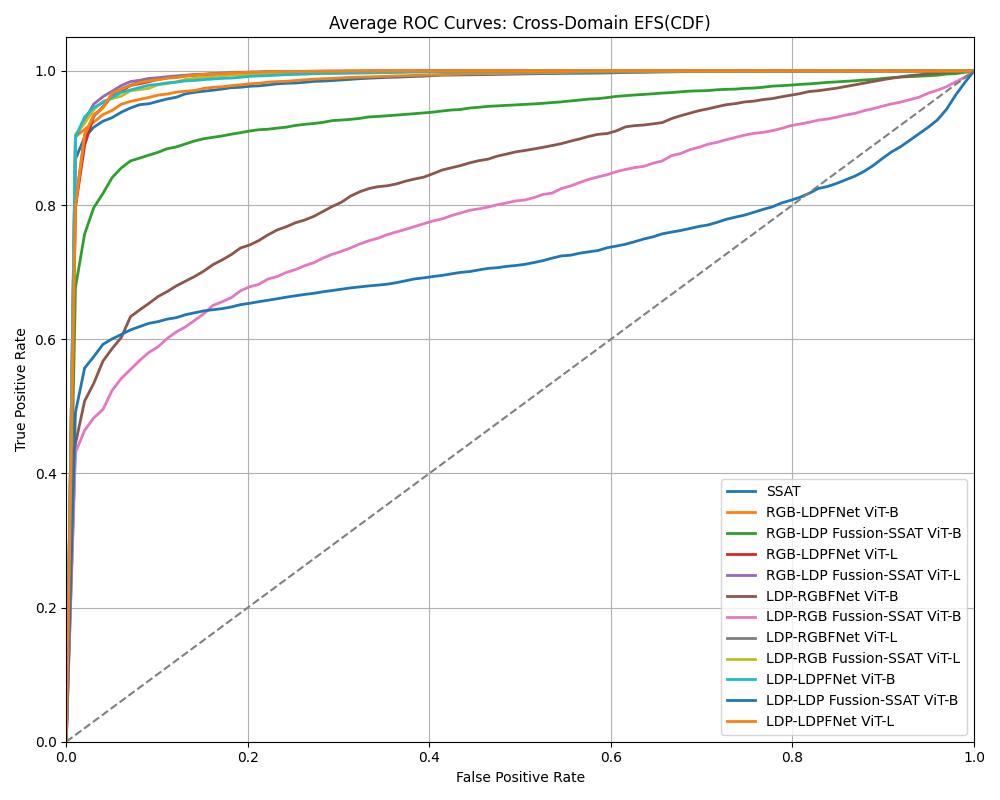}
        \caption{EFS (CDF)}
        \label{fig:cd3}
    \end{subfigure}
    \caption{ROCs of cross-domain evaluations on the DF40 dataset.}
    \label{fig:cross_domain_roc}
\end{figure}

Our proposed model Fusion-SSAT, demonstrates superior generalizability compared to the detectors listed in Table~\ref{tab:deepfakebench}. Among all the above mentioned state-of-the-art detectors, UCF exhibits the highest average at 0.9527\%, however our technique, Fusion-SSAT, achieves an AUC score of 0.9613\%, demonstrating a 2\% above the existing state-of-the-art in within domain evaluation.  Our model Fusion-SSAT showed superior performance in individual evaluations of each forgery method, specifically in FF-c23, FF-DF, FF-F2F, FF-FS, and FF-NT.  In identifying compressed deepfake FF-c40, our model Fusion-SSAT exhibits a performance decrease of 0.005\%. In Cross-Domain Evaluation, each dataset demonstrates a 5\% improvement, and on average, we achieved an 8\% higher AUC score compared to existing state-of-the-art approaches, indicating that our model, Fusion-SSAT, exhibits superior generalizability to on unseen deepfake manipulation techniques. The ROCs are in Fig.~\ref{fig:roc-evaluations}.

\subsubsection{Experimental results on Cross-Forgery and Domain}
Table~\ref{tab:merged} summaries the experimental results of our proposed model trained on the DF40 ff domain, which includes FS(FF), FR(FF), and EFS(FF), and evaluated on within-forgery, cross-forgery, and cross-domain data.  Column-1 specifies the forgery technique utilised for model training, column-2 defines the model employed for evaluation, and column-3 presents the experimental results of the DF40 test set (FF and CDF domains).

\noindent\textbf{Cross-Forgery Evaluation:} Our proposed model, Fusion-SSAT, outperformed all existing models with an average AUC of 0.970\% on FS (FF), 0.958\% on FR (FF), and 0.999\% on EFS (FF). In direct comparisons, our model shows a significant upward trend in AUC: an improvement of ~10\% when trained on FF (FF), ~12\% when trained on FR (FF), and up to ~30\% when trained on EFS (FF).

While all state-of-the-art models exhibit poor performance on the EFS test data, our models demonstrate better generalizability on both FS and FR test sets. When trained on the EFS (FF) training set, SOTA models show a performance drop of up to ~50\% across all test scenarios. In contrast, our proposed method consistently outperforms them, achieving an AUC of 0.99\% on FS (FF) and FR (FF) test sets. The ROCs for these experiments can be found in Fig. \ref{fig:cross_forgery_roc-DF}.

\noindent\textbf{Cross-Domain evaluation:} Most of the current SOTA deepfake outperforms when trained and tested on FF domain witnessed in Table~\ref{tab:merged}, but exhibit poor performance when it is tested on cross-domain data (CDF) as shown in Table~\ref{tab:merged}.

Our proposed model, Fusion-SSAT outperformed all existing SOTA models listed by average AUC of 0.986\% on FS(CDF), 0.981\% on FR(CDF), 0.991\% on EFS(CDF) CDF Test set. Even in one to one cross-domain test, all SOTA models' performance is only up to 70\% AUC, while our proposed models' AUC is not less than 89\%, which shows a significant and better generalizability capability when compare to the existing deepfake detectors. In the FR (CDF) test, the majority of current methodologies exhibit diminished performance, with average AUCs falling below 0.60\%.  On the other hand, our models attain significantly higher average AUCs of 0.97\% and 0.98\%. In all CDF domain tests where current methods demonstrate poor performance, particularly in the FR (FF) scenario where average AUCs frequently fall below 0.60, our models demonstrate exceptional reliability, with AUCs consistently above 0.940 across all test sets. The ROCs for these experiments can be found in Fig. \ref{fig:cross_domain_roc}. 

Fusion-SSAT consistently outperforms all other models across all testing scenarios, achieving the highest average AUC scores.   This demonstrates its remarkable generalisation capability across diverse types of forgeries and areas.   These findings confirm the effectiveness of combining texture-based features with global features to improve cross-domain deepfake detection.

\subsection{Ablation Study}
As part of the ablation study, we evaluate the impact of different feature fusion strategies, including LBP-LBP, LDP-LDP, and LDP-RGB. The results, summarised in Table \ref{tab:ab1}, \ref{tab:ab2} reveal that relying solely on local texture features (LBP-LBP, LDP-LDP) yields suboptimal performance. Notably, combining local descriptors with global representations (LDP-RGB) leads to more discriminative features and improved overall performance.
\vspace{-2mm}

\begin{table*}
\caption{\em \small In-domain and cross-domain evaluation with fusion of LDP-LDP and LDP-RGB features trained on FF++\cite{ff++} dataset    
}
\renewcommand\arraystretch{1.8}
\setlength\tabcolsep{2.5pt}
\centering
\vspace{1mm}
\resizebox{\linewidth}{!}{
\begin{tabular}{>{\centering\arraybackslash}m{3.7cm}>{\centering\arraybackslash}m{2.8cm}*{6}{>{\centering\arraybackslash}p{1.2cm}}*{1}{>{\centering\arraybackslash}p{1.0cm}}*{5}{>{\centering\arraybackslash}p{1.2cm}}>{\centering\arraybackslash}p{1.0cm}}
\toprule
\multirow{2}{*}{\centering\textbf{Detector}} & \multirow{2}{*}{\centering\textbf{Backbone}} & \multicolumn{7}{c}{\textbf{ In-Domain Evaluation}} & \multicolumn{6}{c}{\textbf{Cross Domain Evaluation}}\\
\cmidrule(lr){3-9} \cmidrule(lr){10-15}
&& FF-c23 & FF-c40 & FF-DF & FF-F2F & FF-FS & FF-NT & Avg. & CDFv1 & CDFv2 & DFD & Fsh & UADFV & Avg. \\
\midrule
LDP-RGB Fusion-SSAT & VideoMAE/ViT-B & 0.8833 & 0.7694 & 0.9899 & 0.8163 & 0.9467 & 0.7814 & 0.8645 & 0.5963 & 0.7071 & 0.6500 & 0.9274 & 0.9115 & 0.7585 \\

LDP-RGB Fusion-SSAT & VideoMAE/ViT-L & \textbf{ {0.9866}} & 0.7731 & 0.9997 & 0.9887 & 0.9973 & \textbf{{ 0.9605}} & \underline{0.9510} & \textbf{{ 0.8748}} & \textbf{{ 0.8254 }}& \textbf{{ 0.8557}} & \textbf{{ 0.9947}} & \textbf{{ 0.9540}} & \underline{0.9009} \\

LDP-LDP Fusion-SSAT & VideoMAE/ViT-B & 0.7109 & 0.6696 & 0.9242 & 0.5941 & 0.7051 & 0.6209 & 0.7041 & 0.5309 & 0.754 & 0.6157 & 0.7482 & 0.8426 & 0.6983 \\

LDP-LDP Fusion-SSAT & VideoMAE/ViT-L & 0.9846 & \textbf{{0.7786}} & \textbf{{0.9999}} & \textbf{ {0.9849}} & \textbf{ {0.9966}}  & 0.9581 & 0.9505 & 0.8382 & 0.8158 & 0.8058 & 0.9937 & 0.9281 & 0.8763 \\
\midrule

LBP-RGB Fusion-SSAT & VideoMAE/ViT-B & 0.8243 & \textbf{0.7023} & 0.9689 & 0.8262 & \textbf{0.9052} & 0.7325 & 0.8266 & 0.5262 & 0.7001 & 0.6234 & \textbf{0.9532} & 0.9013 & 0.7408 \\

LBP-RGB Fusion-SSAT & VideoMAE/ViT-L & \textbf{0.9123} & 0.6512 & \textbf{0.9781} & \textbf{0.9345} & 0.8469 & 0.7934 & 0.8527 & \textbf{0.7876} & \textbf{0.8011} & \textbf{0.7350} & 0.8692 & \textbf{0.9021} & \underline{0.8190} \\

LBP-LBP Fusion-SSAT & VideoMAE/ViT-B & 0.5311 & 0.6087 & 0.7420 & 0.5803 & 0.6234 & 0.5698 & 0.6092 & 0.4882 & 0.5405 & 0.6120 & 0.5931 & 0.6598 & 0.5787 \\

LBP-LBP Fusion-SSAT & VideoMAE/ViT-L & 0.8243 & 0.6625 & 0.9711 & 0.8442 & 0.9107 & \textbf{0.7981} & \underline{0.8351} & 0.7821 & 0.7940 & 0.7011 & 0.8250 & 0.8792 & 0.7963 \\

\bottomrule
\end{tabular}
}
\label{tab:ab1}
\vspace{2mm}
\end{table*}

\begin{table*}
\tiny
\centering
\centering
\scriptsize
\setlength{\tabcolsep}{9.8pt} 
\renewcommand{\arraystretch}{1.2} 
\caption{\em \small Cross-forgery and Cross-domain evaluation with fusion of LDP-LDP and LDP-RGB features trained on DF40 dataset}
\label{tab:ab2}
\begin{tabular}{c|c|cccc|cccc}
\hline
\multirow{2}{*}{\textbf{Training Set}} & \multirow{2}{*}{\textbf{Model}} & \multicolumn{4}{c|}{\textbf{Testing Set (FF)}} & \multicolumn{4}{c}{\textbf{Testing Set (CDF)}} \\ \cline{3-10}
                              &                        & FS & FR & EFS & Avg. (FF) & FS & FR & EFS & Avg. (CDF) \\ \hline
\multirow{6}{*}{\textbf{FS (FF)} }    
& LDP-RGB Fusion-SSAT ViT-B& 0.948  & 0.906 & \textbf{{0.955}} & 0.936 & 0.940 & 0.826 & \textbf{{0.995}} & 0.920 \\

& LDP-RGB Fusion-SSAT ViT-L & 0.989 & 0.980 & 0.932 & 0.967 & 0.995 & 0.962 & 0.988 & 0.982 \\

& LDP-LDP Fusion-SSAT ViT-B& 0.687  & 0.620 & 0.721 & 0.676 & 0.774 & 0.654 & 0.918 & 0.782 \\

& LDP-LDP Fusion-SSAT ViT-L & \textbf{{0.991}} & \textbf{{0.985}} & 0.963 & \underline{0.979} & \textbf{{ 0.997}} & \textbf{{0.966}} & 0.991 & \underline{0.985} \\ \hline

\multirow{5}{*}{\textbf{FR (FF)} }    
& LDP-RGB Fusion-SSAT ViT-B & 0.834 & 0.895 & 0.799 & 0.842 & 0.865 & 0.896 & 0.929 & 0.897 \\

& LDP-RGB Fusion-SSAT ViT-L & 0.983 & 0.990 & 0.917 & 0.963 & \textbf{{0.991}} & \textbf{{ 0.994 }}& 0.960 & 0.982 \\

& LDP-LDP Fusion-SSAT ViT-B& 0.609 & 0.760 & 0.70 & 0.689 & 0.654 & 0.726 & 0.827 & 0.736 \\

& LDP-LDP Fusion-SSAT ViT-L & \textbf{{0.992}} & \textbf{{0.995}} & \textbf{{0.985}} & \underline{0.990} & 0.990 & 0.992 & \textbf{{0.984}} & \underline{0.989}\\ \hline

\multirow{5}{*}{\textbf{EFS (FF)} }   
& LDP-RGB Fusion-SSAT ViT-B & 0.940 & 0.836 & 1.0 & 0.925 & 0.767 & 0.600 & 1.0 & 0.789 \\

& LDP-RGB Fusion-SSAT ViT-L & \textbf{{0.999}} & \textbf{{0.998}} & \textbf{{1.0}} & 0.999 & \textbf{{0.997}} & \textbf{{0.981}} & \textbf{{1.0}} & \underline{0.993} \\

& LDP-LDP Fusion-SSAT ViT-B & 0.929 & 0.786 & 1.0 & 0.905 & 0.69 & 0.496 & 0.999 & 0.728 \\

& LDP-LDP Fusion-SSAT ViT-L & 0.999 & 0.998 & \textbf{{1.0}} & \underline{0.999} & 0.997 & 0.976 & 0.999 & 0.991 \\ \hline

\end{tabular}
\end{table*}

\section{Conclusion}

In this work we introduced a novel approach Fusion-SSAT which fuses the feature representation from self-supervised auxiliary task with the primary task. We utilized masked LDP local descriptor as input to the SSAT task and try to reconstruct the RGB patches which makes SSL task harder to learn the local patterns rather than relaying on superficial features. This increased the performance of the primary task to better generalize on unseen manipulations or cross-dataset evaluation. Our experimental results showed a better generalization on large scale DF40 dataset which includes latest deepfake generation techniques than state-of-the-art deepfake detectors.        
{\small
\bibliographystyle{ieee}
\bibliography{egbib}
}
\vspace{-15mm}
\begin{IEEEbiography}[{\includegraphics[width=1in,height=1.25in,clip,keepaspectratio]{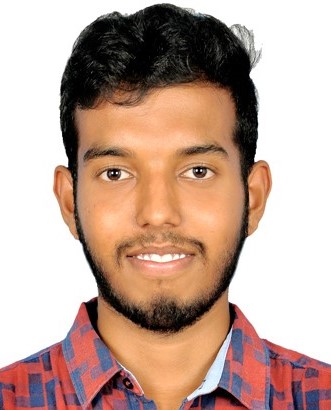}}]{Shukesh Reddy} is currently a PhD scholar in the Department of Computer Science and Information Systems at BITS Pilani, Hyderabad Campus. He earned his Bachelor's degree in Computer Science in 2023. Before pursuing his PhD, he worked as a Software Engineer, leading the development of an IoT platform and IoT development kit. His research interests focus on learning representations of human faces, with his current work centred on advancing face forgery detection techniques.      
\end{IEEEbiography}

\vspace{-15mm}
\begin{IEEEbiography}
[{\includegraphics[width=1in,height=1.25in,clip,keepaspectratio]{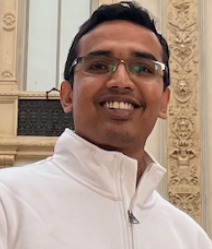}}]{Srijan Das} I am an Assistant Professor in the Department of Computer Science at the University of North Carolina at Charlotte. At UNC Charlotte, I am working on Video Representation Learning, and Robotic Vision. I am a member of the AI4Health Center and one of the founding members of the Charlotte Machine Learning Lab (CharMLab) at UNC Charlotte.

\end{IEEEbiography}

\vspace{-15mm}
\begin{IEEEbiography}[{\includegraphics[width=1in,height=1.25in,clip,keepaspectratio]{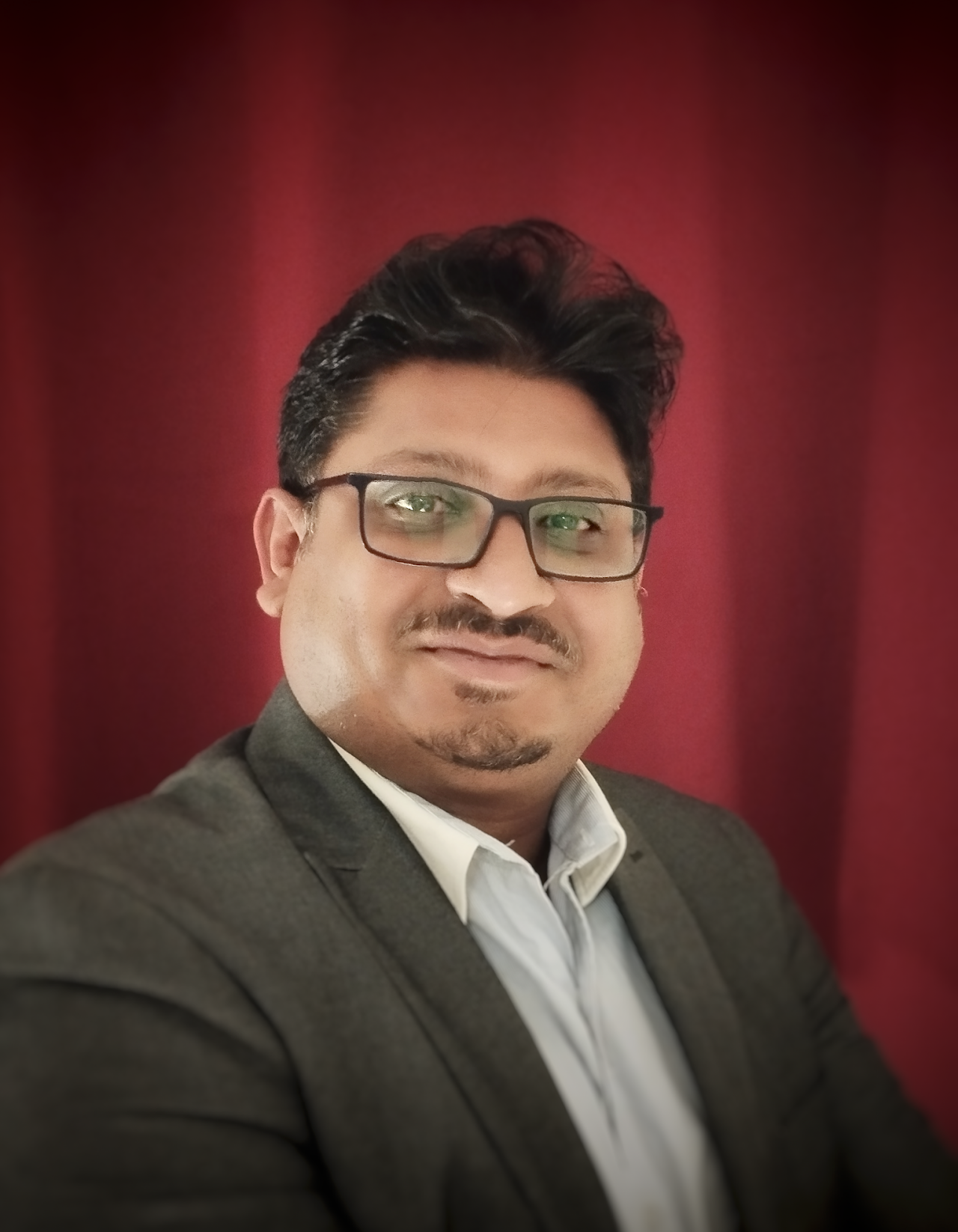}}]{Abhijit Das} is an assistant professor at BITS Pilani Hyderabad. Previously, he worked as a Post-Doc Researcher at Inria Sophia Antipolis– Méditerranée, France. He has completed his PhD from the School of Information and Communication Technology, Griffith University, Australia. He is an accomplished machine learning and computer vision researcher with more than 15 years of research and teaching experience. He is presently pursuing an investigation on learning representations and human analysis employing facial and corporeal-based visual features.
\end{IEEEbiography}

\vspace{-15mm}
\end{document}